\PassOptionsToPackage{T2A,T1}{fontenc}
\PassOptionsToPackage{russian,main=english}{babel}
\PassOptionsToPackage{utf8}{inputenc}
\documentclass[9pt,twocolumn,twoside]{pnas-new}
\setboolean{displaywatermark}{false} 

\usepackage{tipa}
\usepackage{amsmath}
\usepackage{csquotes}

\templatetype{pnasbriefreport} 

\title{Large language models predict human sensory judgments across six modalities}

\author[a,1]{Raja Marjieh}
\author[b]{Ilia Sucholutsky} 
\author[c]{Pol van Rijn}
\author[c,2]{Nori Jacoby}
\author[a,b,2]{Thomas L. Griffiths}

\affil[a]{Department of Psychology, Princeton University, USA}
\affil[b]{Department of Computer Science, Princeton University, USA}
\affil[c]{Max Planck Institute for Empirical Aesthetics, Germany}

\leadauthor{Marjieh}

\authorcontributions{All authors contributed to the design of the experiments and editing the manuscript. Authors RM, NJ, and TLG conceptualized the project. Author IS conducted the LLM experiments. Authors RM and PVR conducted the human experiments. Authors RM, IS, PVR, and NJ analyzed the results.}
\correspondingauthor{\textsuperscript{1}To whom correspondence should be addressed. E-mail: raja.marjieh@princeton.edu}
\authordeclaration{The authors declare no competing interests.}
\equalauthors{\textsuperscript{2}NJ contributed equally to this work with TLG.}


\begin{abstract}
Determining the extent to which the perceptual world can be recovered from language is a longstanding problem in philosophy and cognitive science. We show that state-of-the-art large language models can unlock new insights into this problem by providing a lower bound on the amount of perceptual information that can be extracted from language. Specifically, we elicit pairwise similarity judgments from GPT models across six psychophysical datasets. We show that the judgments are significantly correlated with human data across all domains, recovering well-known representations like the color wheel and pitch spiral. Surprisingly, we find that a model (GPT-4) co-trained on vision and language does not necessarily lead to improvements specific to the visual modality. To study the influence of specific languages on perception, we also apply the models to a multilingual color-naming task. We find that GPT-4 replicates cross-linguistic variation in English and Russian illuminating the interaction of language and perception.
\end{abstract}

\dates{This manuscript was compiled on \today}

\begin{document}

\maketitle
\ifthenelse{\boolean{shortarticle}}{\ifthenelse{\boolean{singlecolumn}}{\abscontentformatted}{\abscontent}}{}

\begin{figure*}[ht]
\centering
\includegraphics[width=\linewidth]{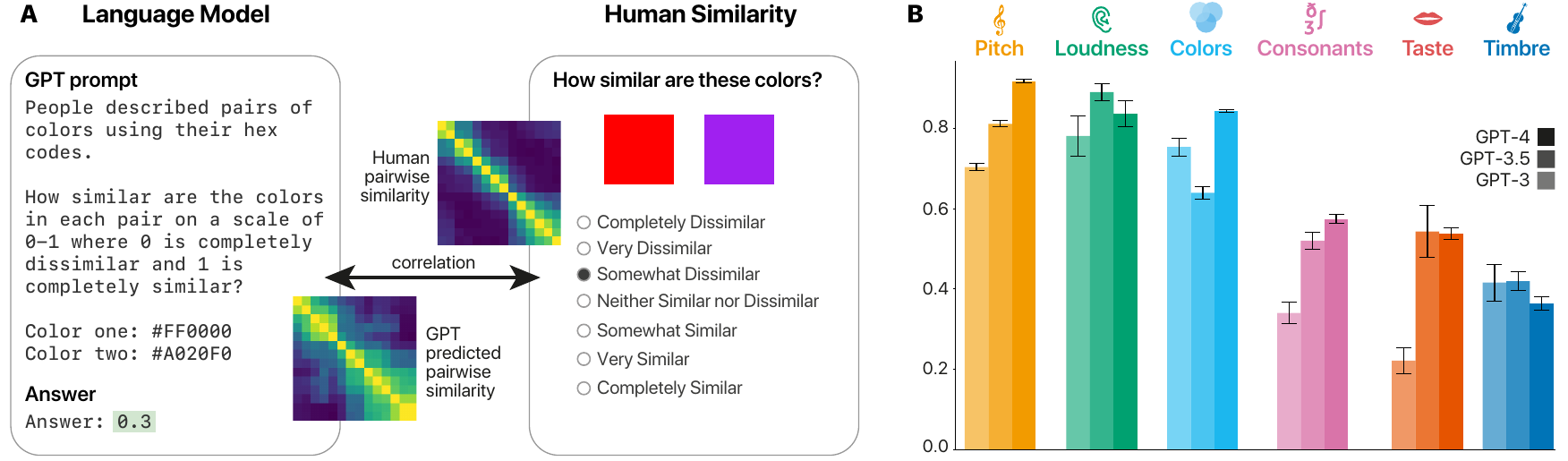}
\caption{\textbf{A.} Schematic of the LLM-based and human similarity judgment elicitation paradigms. \textbf{B.} Correlations between models and human data across six perceptual modalities, namely, pitch, loudness, colors, consonants, taste, and timbre (Pearson $r$; 95\% CIs).}
\label{fig:overview}
\end{figure*}

\dropcap{I}magine that you were chosen to be part of an expedition aimed at studying a newly discovered alien species on a distant planet. Your task is to understand the perceptual system of that species. You arrive at the planet and, to your dismay, discover that its inhabitants are long departed and the only thing they left behind is a huge archive of text. How much of the perceptual world of that species can you recover based on text alone? Versions of this question have occupied philosophers for centuries \cite{hume1938abstract,locke1847essay}, and decades of psychological research are beginning to provide glimpses into the rich perceptual content of language and its influence on perception \cite{goldstone2002using,regier2007color,regier2009language, dolscheid2013thickness, zaslavsky2018efficient,kim2019knowledge,lupyan2020effects,kim2021shared, kawakita2023my}.

But how can the amount of information that language provides about perception be quantified? Here we propose to do so by eliciting psychophysical judgments from large language models (LLMs) such as GPT-3 and its recent ``ChatGPT'' variants GPT-3.5 and GPT-4 \cite{brown2020language,openai2023gpt4}. These models are trained on massive text corpora reflecting a substantial chunk of human language and can be queried in a way that is analogous to humans. \\
\indent Recent research shows that LLMs can be used to study various aspects of cognition \cite{srivastava2022beyond}, including, language processing in the brain \cite{goldstein2022shared,kumar2022reconstructing,tikochinski2023perspective}, perception \cite{abdou2021can,siedenburg2023does,zhang2022visual}, cross-modal alignment \cite{marjieh2022words}, and morality \cite{dillion2023can,ganguli2023capacity}. Here, we use the fact that they are trained on large amounts of language to gain insight into the classic problem of the relationship between language and perception: these models provide a lower bound on the amount of information about perceptual experience that can be extracted from language alone.

We explored this empirically across six modalities, namely, pitch, loudness, colors, consonants, taste, and timbre. Given a stimulus space (e.g., colors) and its stimulus specification (e.g., wavelengths or their corresponding hex-codes) we elicit pairwise similarity judgments in a direct analogy to the widespread paradigm of similarity in cognitive science \cite{shepard1980multidimensional} using a carefully crafted prompt that is given to the model to complete (Figure \ref{fig:overview}A; see Methods). Importantly, whereas four of the modalities were based on classical results from the literature (colors~\cite{ekman1954dimensions}, loudness~\cite{kornbrot1978theoretical}, timbre~\cite{esling2018generative} and taste~\cite{hettinger1999study}), two human datasets (pitch and vocal consonants) were novel and thus were not part of the training set of the models. 

It is worthwhile to note that, unlike the other models, GPT-4 was trained in a multi-modal approach, enabling it to access both written text (similar to the other two variants) and images. This allowed us to examine if the additional sensory information resulted in enhanced performance in the color modality relative to the other domains. Moreover, to further interrogate whether the LLMs' sensory representation was language-dependent, we tested whether they would behave differently in the presence of the same sensory information (a color hex-code), but respond in different languages. To that end, we conducted a color-naming task using a paradigm similar to that of \cite{berlin1991basic} and the World Color Survey \cite{kay2009world,lindsey2014color}, and constructed human and GPT-3, GPT-3.5, and GPT-4 color-naming maps in both English and Russian.

\section*{Results}
\subsection*{Similarity study} 
For each dataset, we designed a tailored prompt template that could be filled in with in-context examples and the pair of target stimuli for which we would want the LLM to produce a similarity rating (see Methods and SI for full specification of the prompts and datasets). Across all domains, we elicited 10 ratings per pair of stimuli from each of the GPT models and then constructed aggregate similarity ratings by averaging. We then evaluated the resulting scores by correlating them with human data. The Pearson correlation coefficients between human data and model predictions are shown in Figure~\ref{fig:overview}B (See SI for details regarding computing the correlations and CIs). We see that across all domains, the correlations were significant, and were particularly high for pitch ($r=.92$, 95\% CI $[.91,.92]$ for GPT-4), loudness ($r=.89$, 95\% CI $[.87,.91]$ for GPT-3.5), and colors ($r=.89$, 95\% CI $[.87,.91]$ for GPT-4) (and $>.6$ for all models), followed by moderate but highly significant correlations for consonants ($r=.57$, 95\% CI $[.56,.59]$ for GPT-4), taste ($r=.54$, 95\% CI $[.48,.61]$ for GPT-3.5) and timbre ($r=.42$, 95\% CI $[.40,.44]$ for GPT-3.5). For the two modalities for which we collected data, we could compare model performance to the inter-rater split-half reliability (IRR). The IRRs for pitch and consonants were $r=.90$ (95\% CI $[.87,.92]$) and $r=.46$ (95\% CI $[.36,.56]$), respectively, suggesting that the performance of GPT-4 is on par with human performance.\\
\indent We also note that in five out of the six domains, GPT-4 was among the top two models. Interestingly, the improvement relative to the average performance of the other models happened across all modalities with the exception of timbre and loudness, and was not restricted or particularly large for the domain of colors (compare e.g. $\Delta r = .15$ 95\% CI $[.13,.16]$ for colors vs. $\Delta r = .16$ 95\% CI $[.15,.17]$ for pitch, $\Delta r = .16$ for taste 95\% CI $[.12,.19]$, and $\Delta r = .14$ 95\% CI $[.12,.16]$ for consonants) suggesting that this improvement is driven by richer textual training in GPT-4 rather than the possibility of its inclusion of images in its training set as is currently being debated \cite{abdou2021can}. \\
\begin{figure*}[ht!]
\centering
\includegraphics[width=\linewidth]{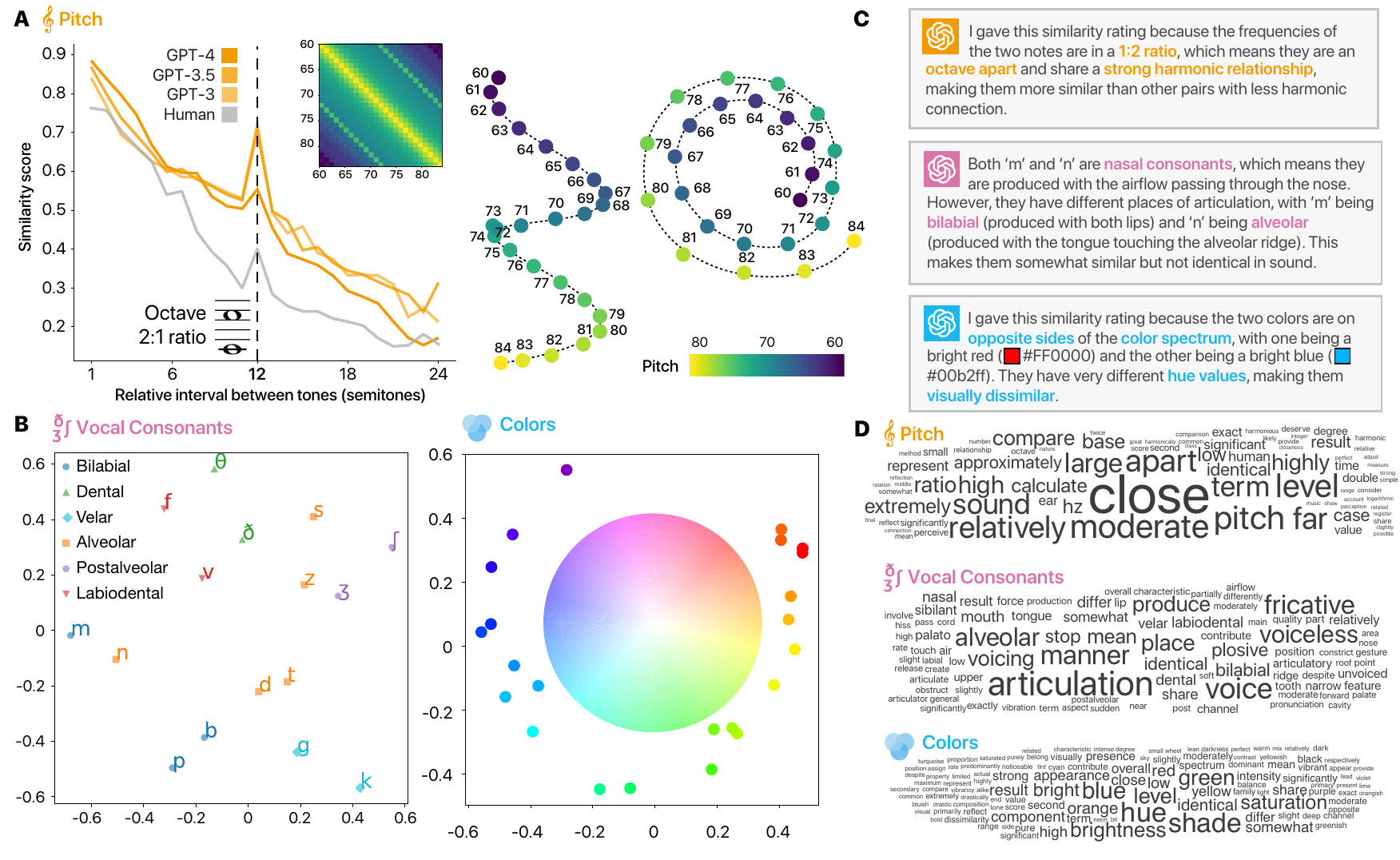}
\caption{\textbf{A.} Human and LLM similarity marginals and an example GPT-3 corresponding similarity matrix and its three-dimensional MDS solution for pitch. \textbf{B.} MDS solutions for vocal consonants and colors for GPT-4 similarity matrices. To illustrate the structure of the results, we highlighted consonants with the same place of articulation in the vocal tract with the same shape and color, and added a rotated HSV color wheel for the color MDS. \textbf{C.} Example GPT-4 explanations for similarity judgment scores. \textbf{D.} Word clouds for GPT-4 explanations in the domain of pitch, vocal consonants, and colors.}
\label{fig:pitch_and_consonants}
\end{figure*}
\indent Next, to get a finer picture of the LLM-based judgments and see to what extent they reflect human representations, we performed the following analyses. Starting from the domain of pitch, we wanted to see to what extent the LLM data captures a well-known psychological phenomenon that Western listeners tend to associate particular musical intervals or ratios of frequencies (such as the octave or 2:1 frequency ratio) with  enhanced similarity \cite{shepard1982geometrical}.  
To test this we computed the average similarity score over groups of pitch pairs that are separated by the same fixed interval (i.e., the same frequency ratios). Figure \ref{fig:pitch_and_consonants}A shows the resulting average similarity per interval for the models and humans along with an example corresponding smoothed similarity matrix for GPT-3 (smoothing was done by averaging the raw similarity matrix over its sub-diagonals). We can see that apart from the decay as a function of separation (i.e., tones that are far apart in log frequency are perceived as increasingly dissimilar) there is a clear spike precisely at 12 semitones (octave), consistent with the aforementioned phenomenon of ``octave equivalence''~\cite{jacoby2019universal}. Moreover, applying multi-dimensional scaling (MDS) \cite{shepard1980multidimensional} to the smoothed similarity matrix whereby the different stimuli are mapped into points in a Euclidean space (also known as ``psychological space'') such that similar stimuli are mapped to nearby points reveals a clear helical structure with twists that correspond to precisely 12 semitone separations (i.e., octaves) recovering the pitch spiral representation (Figure \ref{fig:pitch_and_consonants}A). Likewise, applying MDS to the domains of consonants and colors (Figure~\ref{fig:pitch_and_consonants}B) reveals highly interpretable representations, namely, the familiar color wheel and a production-based representation for consonants.\footnote{An interactive visualization of all human similarity matrices and their corresponding LLM counterparts, as well as their two-dimensional MDS solutions is available at: \url{https://computational-audition.github.io/LLM-psychophysics/all-modalities.html}} As an additional test, we asked GPT-4 to provide explanations for the judgments it made (Figure~\ref{fig:pitch_and_consonants}C-D), and remarkably, the model resorted to explanations involving the octave, ratios, and harmonic relations for pitch, places of articulation in the vocal tract for consonants, and hue, brightness and color spectra for colors, consistent with the MDS solutions.

\subsection*{Color naming study}
The results so far suggest that LLMs can use textual information to form perceptual representations. If this is indeed the case, we hypothesized that representations in LLMs using different languages could be different even in the presence of identical input. This would be consistent with cross-cultural differences in language and perception that were observed in humans~\cite{kay2009world}. To test this, we propose for the first time to test LLMs on an explicit naming task proposed by the seminal work of \cite{berlin1991basic} and further explored across cultures around the globe \cite{kay2009world,lindsey2014color}. 

\begin{figure*}
\centering
\includegraphics[width=\linewidth]{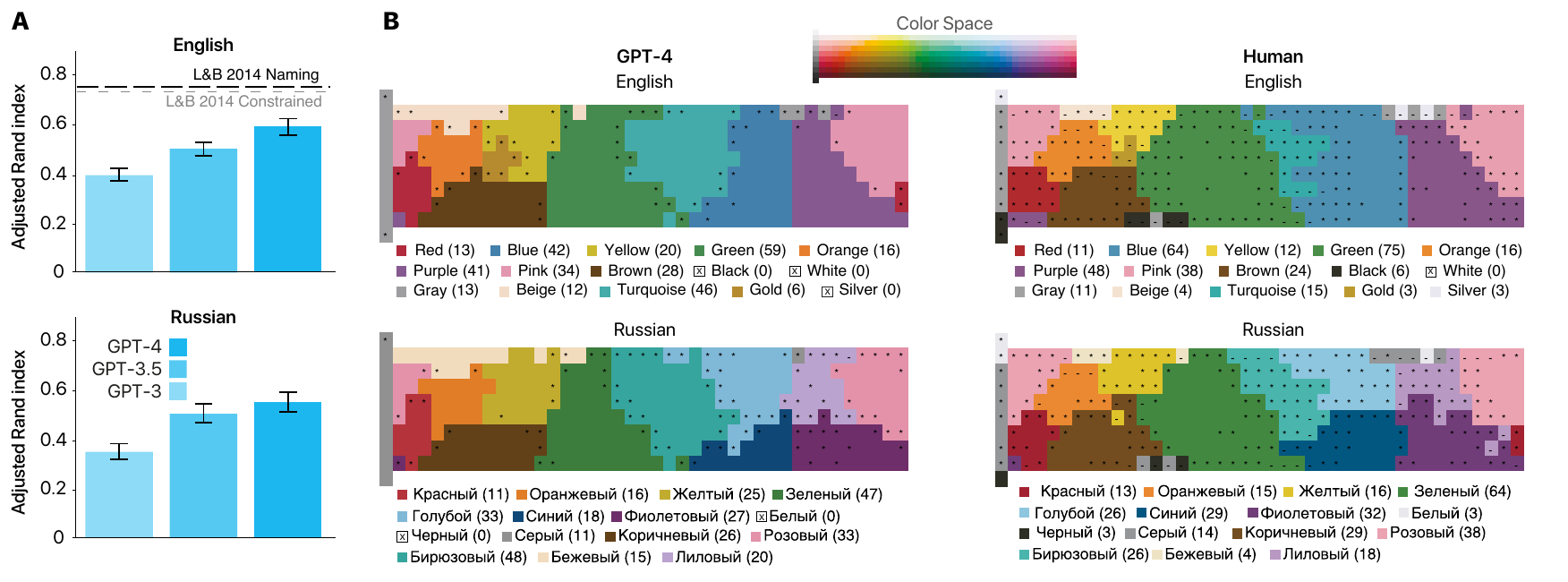}
\caption{Color naming experiment using 330 Munsell colors from the World Color Survey (top, color space). \textbf{A.} Adjusted Rand index illustrating the alignment between human and LLM experiments (95\% CIs). The dashed lines for English represent lab-based free naming and forced-choice naming experiments collected by Lindsey and Brown \cite{lindsey2014color} (data reproduced with permission). \textbf{B.} Data comparison between humans and LLMs in Russian and English. Participants and LLMs were shown colors and were asked to choose from the same 15-color list. The count of chosen colors for each option is given in parentheses. The color of a response cluster in the maps represents its average color.}
\label{fig:color}
\end{figure*}

We thus tested whether LLMs would yield different naming patterns of color hex codes depending on the language of the prompt used to elicit those names (see Methods; Figure~\ref{fig:color}). Specifically, we presented both humans and LLMs with different colors and asked them to perform a forced-choice naming task by selecting from a pre-specified list of 15 color names (see Methods). We specifically focused on English and Russian as test cases, since Russian speakers are documented to use richer vocabulary to describe what English speakers would otherwise describe as blue and purple~\cite{berlin1991basic,paramei2018online,winawer2007russian}. We collected data from 103 native English speakers and 51 native Russian speakers and compared them against LLMs performing the same task, and to the in-lab data of \cite{lindsey2014color} as an additional baseline.\footnote{The human color naming data that we collected can be explored via: \url{https://computational-audition.github.io/LLM-psychophysics/color.html}}

The results are shown in Figure~\ref{fig:color}. Our first finding was that GPT-4 in both English and Russian were more human-like than the other variants when compared with an adjusted Rand index (see Figure~\ref{fig:color}A; English: GPT-4 $0.59$ 95\% CI $[0.56,0.63]$, GPT-3.5, $0.50$ 95\% CI $[0.46,0.52]$, GPT-3 $0.39$ 95\% CI $[0.37,0.42]$; Russian: GPT-4 $0.54$ 95\% CI $[0.46,0.54]$, GPT-3.5 $0.50$ 95\% CI $[0.45,0.52]$, GPT-3 $0.35$ 95\% CI $[0.29,0.35]$; see Methods). It is evident from our data, however, that LLMs are still not perfect in predicting human color naming as compared to a separate lab-based experiment conducted by Lindsey and Brown~\cite{lindsey2014color} (dashed line in Figure \ref{fig:color}A top, constrained naming task 0.73 95\% CI [0.65,0.73], free naming task 0.75 95\% CI [0.66,0.75]). Moreover, the naming of GPT-4 colors differs from human data in some important cases, including the color turquoise, which was selected as the dominant color for 46 Munsell colors in GPT4 versus only 15 in human data.  Note, however, that our human English results conducted online and with relative less control over color presentation were highly consistent with \cite{lindsey2014color} that were conducted in the lab and under controlled environment, even though \cite{lindsey2014color} used slightly different paradigms:  free naming  (consistency to our human data: $0.75$ 95\% CI $[0.66,0.75])$ and forced-choice list with a different set of items ($0.73$ 95\% CI $[0.65,0.74]$).

Importantly, however, GPT-4 appears to replicate cross-lingual differences (Figure~\ref{fig:color}B), for example separating Russian blue and purple into distinct categories for lighter and darker areas \cite{winawer2007russian}. Indeed, the color sínij / \foreignlanguage{russian}{Синий} (Blue) was the dominant category for 18 and 29 Munsell colors for GPT-4 and humans, respectively, and the color golubój / \foreignlanguage{russian}{Голубой} (Light-blue) was accordingly the dominant category for 33 and 26 colors. Similarly, the color fiolétovyj / \foreignlanguage{russian}{Фиолетовый} (Violet) was the dominant category for 27 and 32 Munsell colors for GPT-4 and humans, respectively, and the color lilóvyj / \foreignlanguage{russian}{Лиловый} (Lilac) was accordingly the dominant category for 20 and 18 Munsell colors.

\section*{Discussion}
In this work, we showed how recent advances in large language models and, in particular, their flexible prompt engineering capabilities provide an elegant way for extracting clear quantitative psychophysical signals from text corpora. Our results contribute to a variety of thought-provoking issues in perception and language research. In particular, our findings further support recent research suggesting that people that lack direct sensory experiences (e.g., congenitally blind individuals) could still possess a rich understanding of perceptual concepts through language (e.g., colors \cite{kim2021shared}). Likewise, the language-dependence of the color representation of GPT-4 in the naming task suggests that the physical stimulus alone (in our case, the color hex code) is insufficient for explaining the behavioral patterns of the model since the same hex codes triggered different color categorizations in Russian and English. This supports the idea that language is not only shaped by perception but can also reshape it \cite{winawer2007russian}. Finally, the fact that GPT-4 achieved IRR-level performance without any fine-tuning in the newly collected datasets of pitch and vocal consonants, contributes to our understanding of LLMs' ability to mimic human behavior~\cite{openai2023gpt4}.\\
\indent We end by discussing some limitations which point towards future research directions. First, while in our work we experimented with three different types of behavioral data (similarity, explanations, and naming), there are other perceptual measures that one could consider (for example, odd-one-out triplet judgments \cite{hebart2020revealing}). Future work could explore these measures and interrogate to what extent they too yield a consistent representation. Second, our work is restricted to population-level averages as a leading-order analysis, a follow-up study could look into the natural variability of the LLM judgments and see to what extent they capture individual-level differences in humans. Third, while our work provides some evidence for LLMs' ability to capture cross-cultural differences, it remains to be seen how far this holds for other languages, especially those that are underrepresented \cite{blasi2022over}. Finally, it is important to point out that the very same massive training pipelines that make LLMs a powerful proxy of human language also make them particularly susceptible to inheriting biases \cite{zhuo2023exploring}. Researchers should be particularly cautious when interpreting the patterns of behavior elicited from LLMs and should always benchmark them against genuine human data.\\
\indent To conclude, our work showcases how LLMs can be used to explore the limits of what information about the world can be recovered from language alone, and more broadly, it highlights the potential of combining human and machine experiments in order to understand fundamental questions in cognitive science.
\matmethods{
    \subsection*{GPT prompt elicitation}
    The general structure of the prompt elicitation template for the similarity experiments was: one sentence describing the dataset (e.g., \textit{``people described pairs of colors using their hex code.''}), one sentence describing the similarity rating task and scale (e.g., \textit{``how similar are the colors in each pair on a scale of 0-1 where 0 is completely dissimilar and 1 is completely similar?''}), three sets of three lines each corresponding to two stimuli and their actual similarity rating taken from behavioral experiments which serve as few-shot in-context examples, and an additional set of three lines corresponding to the pair of target stimuli and an associated empty rating field for the model to fill in (\textit{``Color one: \#FF0000. Color two: \#A020F0. Rating:''}). These were necessary to ensure that the model provided numerical values, and in all cases consisted of three fixed and randomly chosen comparisons so that most of the content was left for the model to produce. For each pair of target stimuli, we elicited ten ratings from each GPT model. Across all repetitions of all pairs of stimuli for a given dataset, we used only the same three in-context examples to ensure that the model is exposed to only a very small fraction of the similarity judgments against which its ratings were compared (see SI for full prompts).
    
    For the color-naming experiments, we first elicited 15 basic color names from GPT-4 using the prompt \textit{``Name 15 basic colors.''} and a temperature of 0 (to get the highest probability answers). We then had GPT-4 name the hex code corresponding to each of the WCS colors using the following prompt: \textit{``Here is a list of 15 basic color names: $<$shuffled basic color list$>$. Which of these names best describes the following color: $<$hex-code$>$? Respond only using the name.''} We repeated this prompt ten times for each WCS color with the basic color list shuffled each time and temperature set to the default 0.7 to elicit ten names per WCS color. We repeated the full procedure with both prompts translated to Russian (GPT-4 also responded to both of these in Russian; see SI for full prompt).
    
    \subsection*{Stimuli}
    The six human similarity datasets we considered come in two flavors -- direct (dis-)similarity ratings and confusion matrices -- and from two sources -- previous psychological studies from the literature and newly collected datasets. Confusion matrices provide an alternative way to compute similarity scores between stimuli by counting the number of times a stimulus $x$ is confused for a stimulus $y$. By normalizing the counts one gets confusion probabilities $p_{xy}$ which can be converted into similarity scores using the formula $s_{xy}=\sqrt{p_{xy}p_{yx}/p_{xx} p_{yy}}$~\cite{shepard1987toward, sims2018efficient}.
    
    \paragraph{Colors} 
    This dataset was taken from \cite{ekman1954dimensions} (also reproduced in \cite{shepard1980multidimensional}) and comprised direct similarity judgments across a set of 14 colors with wavelengths in the range $434-674$ nanometers. We converted wavelengths into RGB using the script at~\url{https://hasanyavuz.ozderya.net/?p=211} and then we used the \texttt{webcolors} Python package to convert into hex codes. To get better coverage of the color wheel in Figure~\ref{fig:pitch_and_consonants}B  we extended the space to 23 color stimuli by interpolating between the original colors in the dataset and eliciting an extended similarity matrix from GPT-4. 
    
    \paragraph{Pitch} 
    This dataset was collected and made publicly available very recently by a subset of the authors in \cite{Marjieh2023.06.13.544763} (see details below). It contains similarity judgments over pairs of 25 harmonic complex tones (10 partials and 3dB/octave roll-off) over a two octave range from C4 (60 MIDI; 261.626 Hz) to C6 (84 MIDI; 1046.502 Hz). The pitch values were separated by 1 semitone steps to account for the fact that pitch perception is logarithmic~\cite{jacoby2019universal} where the mapping between frequencies $f$ in Hertz and pitch $p$ in semitones are given by $p=12 \log_2{f/440} + 69$.
    
    \paragraph{Vocal consonants} 
    This dataset was also collected via an online study (see below) and comprised similarity judgments over 16 recordings of vocal consonants taken from the International Phonetic Association\footnote{\url{https://www.internationalphoneticassociation.org/}}. The vocal consonants considered were \textipa{b} (\textbf{b}ay), \textipa{p} (\textbf{p}ay), \textipa{m} (\textbf{m}ay), \textipa{n} (\textbf{n}o), \textipa{g} (\textbf{g}o), \textipa{k} (ca\textbf{k}e), \textipa{d} (\textbf{d}ie), \textipa{t} (\textbf{t}ie), \textipa{f} (\textbf{f}ee), \textipa{v} (\textbf{v}ow), \textipa{s} (\textbf{s}o), \textipa{T} (\textbf{th}igh), \textipa{D} (\textbf{th}ey), \textipa{Z} (\textbf{J}acques), and \textipa{S} (\textbf{sh}ow). The recordings came from two speakers, one male and one female.
    
    \paragraph{Loudness} 
    We accessed this dataset via \cite{sims2018efficient} which itself takes the data from \cite{kornbrot1978theoretical}. The dataset comes in the form of a confusion matrix over 8 pure tones of different loudness values ranging from 71.1 to 74.6 decibels.
    
    \paragraph{Taste} 
    This dataset was also accessed via \cite{sims2018efficient} and is taken from \cite{hettinger1999study}. The data comes in the form of a confusion matrix over 10 flavors described to participants as salt, salt-substitute, MSG, quinine, acid, sugar, artificial sweetener, salt-sugar, acid-sugar, and quinine-sugar.
    
    \paragraph{Timbre} 
    This dataset was assembled in \cite{esling2018generative} based on 1217 subject's judgments from 5 prior publications. It comprises dissimilarity judgments over 12 instrument timbres: clarinet, saxophone, trumpet, cello, French horn, oboe, flute, English horn, bassoon, trombone, violin, and piano.
    
    \subsection*{Behavioral experiments}
    To collect similarity judgments over pitch and vocal consonants we deployed two online experiments on Amazon Mechanical Turk (AMT). Overall, 55  participants completed the pitch study and 64 participants completed the vocal consonants study. In addition, we collected  color naming data in Russian and British English participants using Prolific\footnote{\url{https://www.prolific.co}}. Overall, we recruited 154 participants of which 103 were UK participants and 51 were Russian participants. Experiments were implemented using PsyNet\footnote{\url{https://psynet.dev/}} and Dallinger\footnote{\url{https://dallinger.readthedocs.io/}}. See additional details in SI.
    
    \subsection*{Code and Data availability}
    All data and code used in this work can be accessed via the following link: \url{https://tinyurl.com/fudaby5p}, with the exception of the \cite{lindsey2014color} color naming dataset as it is only available upon request from the authors. An interactive visualization of two-dimensional MDS spaces for the 6 modalities is available at: \url{https://computational-audition.github.io/LLM-psychophysics/all-modalities.html}. The human color naming data can be interactively explored via: \url{https://computational-audition.github.io/LLM-psychophysics/color.html}.
}

\showmatmethods{} 

\acknow{This research project and related results were made possible with the support of the NOMIS Foundation, and an NSERC fellowship (567554-2022) to IS.}

\showacknow{} 

\bibliography{manuscript}

\appendix
\newpage
\section*{Extended Methods}
\subsection*{Explicit GPT prompts}
GPT prompt elicitation experiments were conducted using the OpenAI Text Completion (for GPT-3) and Chat Completion (for GPT-3.5 and GPT-4) APIs. The temperature was set to 0.7 for similarity judgments and 0 for color naming experiments. The exact prompt formats used are shown below.
\subsubsection*{Similarity judgments}
\paragraph{Color:}

\begin{displayquote}
    \texttt{%
    People described pairs of colors using their hex codes.
    How similar are the two colors in each pair on a scale of 0-1 where 0 is completely dissimilar and 1 is completely similar?
    Respond only with the numerical similarity rating.
    \\
    Color 1: \#ff5700
    Color 2: \#ff9b00
    Rating 0.76
    \\
    Color 1: \#b3ff00
    Color 2: \#00ff61
    Rating: 0.45
    \\
    Color 1: \#FF0000
    Color 2: \#00b2ff
    Rating: 0.02
    \\
    Color 1: $<$hex-code1$>$
    Color 2: $<$hex-code2$>$
    \\
    Rating:
    }
\end{displayquote}

\paragraph{Pitch:} 
\begin{displayquote}
    \texttt{%
    People described pairs of musical notes using their frequencies in hertz.\\
    How similar are the musical notes in each pair on a scale of 0-1 where 0 is completely dissimilar and 1 is completely similar?\\
    \\      
    Note 1: 587.3295358348151 Hz\\
    Note 2: 987.7666025122483 Hz\\
    Rating: 0.46083740655517463\\
    \\                    
    Note 1: 349.2282314330039 Hz\\
    Note 2: 277.1826309768721 Hz\\
    Rating: 0.743838237117938\\
    \\      
    Note 1: 415.3046975799451 Hz\\
    Note 2: 987.7666025122483 Hz\\
    Rating: 0.19874605585261726\\
    \\      
    Note 1: $<$frequency1$>$\\
    Note 2: $<$frequency2$>$\\
    Rating:
    }
\end{displayquote}

\paragraph{Vocal consonants:} 
\begin{displayquote}
    \texttt{%
    People described vocal consonants using the international phonetic alphabet (IPA).\\
    How similar do the vocal consonants in each pair sound on a scale of 0-1 where 0 is completely dissimilar and 1 is completely similar?\\
    Respond only with the numerical similarity rating.\\
    \\
    Vocal Consonant 1: f\\
    Vocal Consonant 2: m\\
    Rating: 0.5\\
    \\    
    Vocal Consonant 1: n\\
    Vocal Consonant 2: }\textipa{Z}\\
    \texttt{
    Rating: 0.40740740740740744\\
    \\
    Vocal Consonant 1:} \textipa{S}\\
    \texttt{Vocal Consonant 2:} \textipa{S}\\
    \texttt{
    Rating: 1.0\\
    \\
    Vocal Consonant 1: $<$consonant1$>$\\
    Vocal Consonant 2: $<$consonant2$>$\\
    Rating:
    }
\end{displayquote}

\paragraph{Loudness:} 
\begin{displayquote}
    \texttt{%
    People described the loudness of pure tones in decibels (dB).\\
    How similar do the pure tones in each pair sound on a scale of 0-1 where 0 is completely dissimilar and 1 is completely similar?\\
        \\
        Pure Tone 1: 72.6 dB\\
        Pure Tone 2: 74.1 dB\\
        Rating: 0.3495324720283043\\
        \\
        Pure Tone 1: 74.6 dB\\
        Pure Tone 2: 73.6 dB\\
        Rating: 0.5055839477695901\\
        \\
        Pure Tone 1: 74.1 dB\\
        Pure Tone 2: 74.1 dB\\
        Rating: 1.0\\
    \\  
    Pure Tone 1: $<$loudness1$>$\\
    Pure Tone 2: $<$loudness2$>$\\
    Rating:
}
\end{displayquote}

\paragraph{Taste:} 
\begin{displayquote}
    \texttt{%
     People described flavors they tasted using words.\\
    How similar are the flavors in each pair on a scale of 0-1 where 0 is completely dissimilar and 1 is completely similar?\\                \\
        Flavor 1: quinine\\
        Flavor 2: artificial sweetener\\
        Rating: 0.0\\
        \\
        Flavor 1: artificial sweetener\\
        Flavor 2: salt\\
        Rating: 0.015433904145892428\\
        \\
        Flavor 1: quinine-sugar\\
        Flavor 2: acid-sugar\\
        Rating: 0.2539115246067999\\
    \\                    
    Flavor 1: $<$flavor1$>$\\
    Flavor 2: $<$flavor2$>$\\
    Rating:
}
\end{displayquote}

\paragraph{Timbre:} 
\begin{displayquote}
    \texttt{%
    People listened to pairs of musical instruments and rated the similarity of their timbre.\\
    How similar is the timbre of the instruments in each pair on a scale of 0-1 where 0 is completely dissimilar and 1 is completely similar?\\
    \\
        Instrument 1: Cello\\
        Instrument 2: Flute\\
        Rating: 0.5604846433040316\\
    \\      
        Instrument 1: Flute\\
        Instrument 2: Clarinet\\
        Rating: 0.270932601836378\\
    \\    
        Instrument 1: Trombone\\
        Instrument 2: Bassoon\\
        Rating: 0.2893895067551666\\
    \\  
    Instrument 1: $<$instrument1$>$\\
    Instrument 2: $<$instrument2$>$\\
    Rating:
    }
\end{displayquote}
\clearpage

\noindent\subsubsection*{Color naming}
\paragraph{Basic color free-elicitation:}\hfill

\noindent English:
\begin{displayquote}
    Name 15 basic colors.
\end{displayquote}

\noindent Russian:
\begin{displayquote}
    \foreignlanguage{russian}{Перечислите 15 основных цветов.}
\end{displayquote}

\paragraph{Color naming elicitation:} \hfill

\noindent English:
\begin{displayquote}
    Here is a list of 15 basic color names: $<$shuffled basic color list$>$.\\
    Which of these names best describes the following color: $<$hex-code$>$?\\
    Respond only using the name.
\end{displayquote}

\noindent Russian:
\begin{displayquote}
\foreignlanguage{russian}{
    Вот список из 15 названий основных цветов: $<$shuffled basic color list$>.$\\
    Какое из названий цветов лучше всего описывает следующий цвет: $<$hex-code$>$?\\
    Отвечайте только названием одного цвета из списка.\\
}
\end{displayquote}

We repeated this prompt ten times for each WCS color with the basic color list shuffled each time and temperature set to the default 0.7 to elicit ten names per WCS color. For each of the ten elicitations per color, if the output was not one of the 15 basic colors we would keep re-querying GPT until it did give a valid color output (GPT-4 is slightly non-deterministic even at temperature 0 due to changes in hardware). If after 10 attempts the response was still invalid, we would return ``error'' as the color (this response is later discarded from the analysis). 

\subsection*{Additional details of behavioral experiments}
\subsubsection*{Similarity Experiments: Participants}
To collect similarity judgments over pitch and vocal consonants we deployed two online experiments on Amazon Mechanical Turk (AMT).\footnote{\url{https://www.mturk.com/}} The recruitment and experimental pipelines were automated using PsyNet~\cite{HarrisonMarjieh2020}, a modern framework for experiment design\footnote{\url{https://psynet.dev/}} and deployment which builds on the Dallinger\footnote{\url{https://dallinger.readthedocs.io/}} platform for recruitment automation. Overall, 55 participants completed the pitch study and 64 participants completed the vocal consonants study. 
Participants were recruited from the United States, were paid \$9-12 USD per hour, and provided informed consent as approved by the Princeton IRB (\#10859) and the Max Planck Ethics Council (\#2021\_42). 

To enhance data quality, participants had to pass a standardized headphone check~\cite{woods2017headphone} that ensures good listening conditions and task comprehension, and were required to have successfully completed at least 3000 tasks on AMT. Upon passing the prescreening stage, participants were randomly assigned to rate the similarity between different pairs of stimuli and provided numerical judgments on a 7-Likert scale ranging from 0 (completely dissimilar) to 6 (completely similar). In the pitch experiment, participants provided an average of 80 judgments, and in the vocal consonants experiment an average of 55 judgments. See SI for explicit instructions.

\subsubsection*{Similarity Experiments: Procedure}
Upon providing informed consent and passing the headphone check, participants received the following instructions. In the case of the pitch experiment: ``In this experiment we are studying how people perceive sounds. In each round you will be presented with two sounds and your task will be to simply judge how similar those sounds are. You will have seven response options, ranging from 0 (`Completely Dissimilar') to 6 (`Completely Similar'). Choose the one you think is most appropriate. You will also have access to a replay button that will allow you to replay the sounds if needed. Note: no prior expertise is required to complete this task, just choose what you intuitively think is the right answer.'' Participants were then informed of an additional small quality bonus ``The quality of your responses will be automatically monitored, and you will receive a bonus at the end of the experiment in proportion to your quality score. The best way to achieve a high score is to concentrate and give each round your best attempt''. While the task is subjective in nature, we used consistency as a proxy for quality by repeating 5 random trials at the end of the experiment and computing the Spearman correlation $s$ between the original responses and their repetitions. The final bonus was computed using the formula $\min(\max(0.0, 0.1s),0.1)$ yielding at most 10 cents. In the main experiment participants were assigned to random stimulus pairs and were instructed to rate their similarity using the following prompt: ``How similar are the pair of sounds you just heard?'' and provided a response on a Likert scale.
The procedure for the vocal consonants similarity experiment was identical up to the specific instructions. Specifically, participants received the following instructions: ``In this experiment we are studying how people perceive the sound of vocal consonants. A consonant is a speech sound that is pronounced by partly blocking air from the vocal tract. For example, the sound of the letter \emph{c} in \emph{cat} is a consonant, and so is \emph{t} but not \emph{a}. Similarly, the sound of the combination \emph{sh} in \emph{sheep} is a consonant, and so is \emph{p} but not \emph{ee}. In general, vowel sounds like those of the letters \emph{a, e, i, o, u} are not consonants.'' The instructions then proceeded: ``In each round you will be presented with two different recordings each including one consonant sound and your task will be to simply judge how similar are the sounds of the two spoken consonants. We are not interested in the vowel sounds nor in the voice height, just the sound of the consonants. You will have seven response options, ranging from 0 (`Completely Dissimilar') to 6 (`Completely Similar'). Choose the one you think is most appropriate. Note: no prior expertise is required to complete this task, just choose what you intuitively think is the right answer.'' Participants were then informed of the quality bonus which was identical to the pitch task, and then rated the similarity between pairs of random consonants based on the following prompt ``How similar is the sound of the consonants pronounced by the two speakers?'' and a Likert scale as before.

\clearpage

\subsubsection*{Similarity Experiments: Model Evaluation}

We quantified model performance in predicting human similarity judgments by computing the Pearson correlation coefficient between the flattened upper triangle of the LLM-based and human-based similarity matrices (to account for the fact that these matrices are symmetric). This approach is similar to representational similarity analysis~\cite{kriegeskorte2008representational}. To compute 95\% confidence intervals, we bootstrapped with replacement over model predictions with 1,000 repetitions and computed for each repetition the average similarity matrix. We then correlated the upper triangles of each of those matrices with human data to produce a list of correlation coefficients on which we computed confidence intervals.

\subsubsection*{Color Naming Experiments: Participants}
To collect the color naming data in Russian and British English participants, we ran online experiments on Prolific\footnote{\url{https://www.prolific.com}}. Overall, we recruited 103 UK participants and 51 Russian participants.  All texts in the interface of the experiment (e.g., buttons, instructions, etc) were presented in the native language of the participant. The Russian texts were first automatically translated using DeepL\footnote{\url{https://www.deepl.com}} and then manually checked and corrected by a native speaker of Russian (author I.S). Participants had to be raised monolingually and to speak the target language as their mother tongue. Each participant was paid 9 GBP per hour and provided informed consent according to an approved protocol (Max Planck Ethics Council \#2021\_42). The experiment was implemented using PsyNet~\cite{HarrisonMarjieh2020}. Each session starts with a free-elicitation task where participants are asked to provide basic colors:
\subsubsection*{Color Naming Experiments: Procedure}

\paragraph{Basic color free-elicitation:} \hfill

\noindent English:
\begin{displayquote}
    Please name at least 8 basic color names.\\
    Press enter after each color name.\\
    Only use lower-case letters.\\
\end{displayquote}

\noindent Russian:
\begin{displayquote}
\foreignlanguage{russian}{%
   Укажите не менее 8 названий основных цветов.\\
   Нажмите клавишу Enter после каждого названия цвета.\\
   Используйте только строчные буквы.\\
}
\end{displayquote}

Participants may only submit color names without spaces, numbers, or special characters and can only submit the page if they have provided at least eight names. The list of obtained colors is highly overlapping with the GPT-4 list, justifying our choice to use GPT-4 as the basis for the word naming task (colors are sorted by their naming frequency). \\

\noindent Top 15 terms in English:
\begin{displayquote}
    ``blue'', ``green'', ``yellow'', ``red'', ``purple'', ``orange'', ``black'', ``pink'', ``white'', ``brown'', ``grey'', ``violet'', ``indigo'', ``turquoise'', ``silver''\\
\end{displayquote}

\noindent Top 15 terms in Russian:
\begin{displayquote}
\foreignlanguage{russian}{%
    ``красный'', ``синий'', ``белый'', ``зеленый'', ``оранжевый'', ``желтый'', ``фиолетовый'', ``черный'', ``голубой'', ``коричневый'', ``розовый'', ``серый'', ``жёлтый'', ``зелёный'', ``чёрный''\\
}
\end{displayquote}

From the top 15 color terms, 11 (English) and 12 (Russian) color terms are overlapping with the list provided by GPT-4. Before the main experiments, participants received the following instructions:\pagebreak

\paragraph{Color naming instructions:} \hfill

\noindent English:
\begin{displayquote}
    During this experiment, you will be presented with a square of a particular color and will be required to select the most suitable color term from a list of options.\\
    Please be aware that some of the colors may be repeated to verify consistency of your choices.\\
    If we detect any inconsistencies in your answers, we may terminate the experiment prematurely.\\
    The best strategy is to answer each question truthfully, as attempting to memorize responses may prove difficult.\\
\end{displayquote}

\noindent Russian:
\begin{displayquote}
\foreignlanguage{russian}{%
    В ходе этого эксперимента вам будет представлен квадрат определенного цвета, и вам нужно будет выбрать наиболее подходящее название цвета из списка  вариантов. \\
    Имейте в виду, что некоторые цвета могут повторяться, чтобы убедиться в согласованности вашего выбора. \\
    Если мы обнаружим какие-либо несоответствия в ваших ответах, мы можем досрочно прекратить эксперимент. \\
    Лучшая стратегия - отвечать на каждый вопрос правдиво, так как попытка  запомнить ответы может оказаться сложной.\\
}
\end{displayquote}

The participants then went through the main experiment:

\paragraph{Color naming task:} \hfill

\noindent English:
\begin{displayquote}
    $<$square of a particular color$>$\\
    You will see below a list of 15 basic color names. Which of these names best describes the color above? \\
    $<$shuffled basic color list presented as buttons$>$\\
\end{displayquote}

\noindent Russian:
\begin{displayquote}
\foreignlanguage{russian}{%
    $<$square of a particular color$>$\\
    Ниже Вы увидите список из 15 основных названий цветов.\\
    Какое из этих названий лучше всего описывает вышеуказанный цвет?\\
    $<$shuffled basic color list presented as buttons$>$\\
}
\end{displayquote}

At the end of the experiment, the participant took a color blindness test ~\cite{clark1924ishihara}. Some participants abandoned the experiment prematurely, but we nevertheless included their responses (42 English participants, 3 Russian participants). Only a fraction of the participants failed the color blindness test (5 of 103 English participants, and  2 of 51 Russian participants). Consistent with the WCS we included all participants including those who failed the color blindness test. In a control analysis, we excluded all color blind individuals and all participants that did not complete the entire session and got nearly identical results (the adjusted Rand index was 0.92 for English experiments and 0.97 for Russian experiments).

\subsubsection*{Color Naming Experiments: Analysis}

For each color, we collected at least 10 responses per LLM variant, and at least 10 forced-choice human selections per color (English mean 19.30 responses, Russian mean 12.17 responses). Consistent with previous literature for each Munsell's color we selected the most frequently reported term. We then presented the dominant colors in Figure~\ref{fig:color}B. To aid visualization we average the RGB values of all colors with the same color term, and presented them as the legend and clustered color in that figure. We also listed per color the number of Munsell's colors that were associated with each dominant color term. Figure~\ref{fig:color}B provides additional information on the degree of agreement for each color.  Colors for which less than 50\% and 90\% of the times the dominant color term was selected were indicated by ``-'' and ``*'', respectively. If the dominant color term was selected more than 90\% of the time, no marking was used.

\subsubsection*{Adjusted Rand index} 
The Rand index~\cite{rand1971objective} is a label-insensitive measure of clustering similarity that instead of relying on specific labels (e.g. ``Blue'') quantifies the similarity between two clustering partitions by counting pairs of items (in our case Munsell colors) that are clustered consistently and dividing them by the overall number of pairs. This allows to compare different clustering schemes when the vocabulary of labels is not aligned (e.g. English and Russian). Formally, we computed: $R=(b+c)/a$; where $b$ is the number of pairs of items that are in the same subset in one clustering and in the same subset in the other, $c$ is the number of pairs of items that are in different subsets in one clustering and in different subsets in the other and $a$ is the total number of pairs.
The Rand index provides high values for two random clusterings, to adjust for this we used the corrected-for-chance version of the Rand index ~\cite{rand1971objective}, which normalizes the raw value by the expected value of the Rand index for random clusterings. Formally, we have $ARI = (RI - RI_{\text{rand}}) / (1 - RI_{\text{rand}})$ where $ARI$ is the adjusted Rand index, $RI$ is the raw Rand index and $RI_{\text{rand}}$ is the expected Rand index for random clusterings. The adjusted Rand index is thus ensured to have a value close to 0.0 for random labeling independently of the number of clusters and samples, exactly 1.0 when the clusterings are identical (up to a permutation), and reaches -0.5 for ``orthogonal clusters'' that are less consistent relative to what is expected by chance. In our case, all values were strictly positive suggesting consistency across languages and experiments. To compute confidence intervals, we created bootstrapped datasets by sampling the responses of each color with replacement and recomputing the dominant selected color name. We then obtained CIs by computing the adjusted Rand index for 1,000 pairs of bootstrapped datasets.

\subsubsection*{Lindsey and Brown dataset}
We compared our experimental data to a dataset by \cite{lindsey2014color}, reproduced with permission by Delwin Lindsey. The data contains two experimental conditions conducted in the lab with the same 51 participants. In the first condition, participants were instructed to provide free naming responses. In the second condition,  participants were instructed to choose from a pre-specified list of 11 color terms: (Green, Blue, Purple, Pink, White, Brown, Orange, Yellow, Red, Black, and Gray). Despite the fact that the Lindsey \& Brown experiment was conducted in the lab (and not online like our experiments) and that the constrained list in our experiment was somewhat different, the results of both experiments were highly consistent with our human English data (Constrained, $ARI=0.73$ 95\% CI $[0.65,0.74]$, free naming $0.75$ 95\% CI $[0.66,0.75]$). In addition to putting an upper bound on the consistency with which the LLM can predict human data (by comparing it with another human experiment), these results prove that despite less control over color presentation compared to the lab, online presentation still provides high-quality color naming data. 

\end{document}